\documentclass{article}
\usepackage{spconf,amsmath,graphicx}
\usepackage{xcolor}
\usepackage[caption=false]{subfig}
\usepackage{url}

\title{FUNQUE: FUSION OF UNIFIED QUALITY EVALUATORS}
\name{Abhinau K. Venkataramanan$^{\star}$ \qquad Cosmin Stejerean$^{\dagger}$ \qquad Alan C. Bovik$^{\star}$\thanks{This research was sponsored by a grant from Facebook Video Infrastructure, and by grant number 2019844 for the National Science Foundation AI Institute for Foundations of Machine Learning (IFML).}}
\address{$^{\star}$ The University of Texas at Austin, Austin, TX 78705 USA \\
        $^{\dagger}$ Meta Platforms, Inc., Menlo Park, CA 94025 USA
    }
\usepackage{scrlayer-scrpage}
\usepackage{blindtext}
\clearpairofpagestyles
 \relax
\begin{document}
%

\maketitle
\begin{abstract}
Fusion-based quality assessment has emerged as a powerful method for developing high-performance quality models from quality models that individually achieve lower performances. A prominent example of such an algorithm is VMAF, which has been widely adopted as an industry standard for video quality prediction along with SSIM. In addition to advancing the state-of-the-art, it is imperative to alleviate the computational burden presented by the use of a heterogeneous set of quality models. In this paper, we unify ``atom'' quality models by computing them on a common transform domain that accounts for the Human Visual System, and we propose FUNQUE, a quality model that fuses unified quality evaluators. We demonstrate that in comparison to the state-of-the-art, FUNQUE offers significant improvements in both correlation against subjective scores and efficiency, due to computation sharing.
\end{abstract}

\section{INTRODUCTION}
\label{sec:introduction}

The share of video in mobile traffic is expected to reach 82\% by the year 2022 \cite{ref:cisco}. Owing to this explosion of videos online, Video Quality Assessment (VQA) has emerged as a key area of research. While the most reliable form of VQA is subjective VQA, where videos are rated by human subjects, practical VQA relies on objective VQA models. In the case of streaming and Video On Demand (VOD) applications, the pristine video is available for use as a reference against which distorted videos may be evaluated. Therefore, Full-Reference (FR) VQA algorithms are of special interest.

The Video Multimethod Assessment Fusion (VMAF) \cite{ref:vmaf} quality model has been widely adopted as an industry standard for quality assessment of compressed videos. In particular, VMAF has found use in the perceptual optimization of encoding recipes, comparison of codecs \cite{ref:cc_hddo, ref:cc_hd}, and to evaluate enhancement/precoding methods \cite{ref:precoding}. The current widely adopted version of the VMAF quality model is VMAF v0.6.1, which uses Spatial Visual Information Fidelity (VIF) \cite{ref:vif} at 4 scales, the Detail Loss Metric (DLM, called ADM in VMAF) \cite{ref:dlm}, and Temporal Difference (TD, called Motion in VMAF), as its ``atom features,'' which are fused using a support vector regressor (SVR).

Notably, the VIF model \cite{ref:vif} was originally defined in the steerable-pyramid wavelet domain \cite{ref:steer}, and has been reformulated using dyadic wavelets \cite{ref:vif_dwt}. Thus, the wavelet domain is the ``natural domain'' for VIF. On the other hand, the spatial VIF model is a version of VIF that has been optimized for speed by omitting the wavelet decomposition, at the cost of a lower correlation against subjective scores. However, since the computation of DLM requires an expensive 4-level wavelet transform, sharing the wavelet decomposition will allow the computation of VIF in its natural domain, while improving computational efficiency. This notion of using a common decomposition to ``unify'' the atom features is a foundational idea of this work.

\begin{table}[b]
    \caption{Design choices considered for FUNQUE}
    \label{tab:design_choices}
    \begin{center}
    \begin{tabular}{|c|p{0.45\linewidth}|}
        \hline
        Design Choice & Options \\
        \hline
        Wavelet & Haar / Db2\\
        Wavelet Levels & 1 - 4 \\
        CSF & Frequency filter / Spatial filter / Li SW / Watson SW \\
        CSF Sharing & Yes / No \\
        SAST & Yes / No \\
        \hline
    \end{tabular}
    \end{center}
\end{table}

\section{RELATED WORK}

Since VMAF is applied on the luma channel and the only temporal feature is the temporal difference of the reference video, attempts at improving VMAF have typically focused on the inclusion of color and temporal features. ColorVMAF \cite{ref:color_vmaf} introduces color information by computing features on the chroma channels, while the use of spatio-temporal features has been explored by Ensemble VMAF \cite{ref:ensemble_vmaf}. The robustness of VMAF to video enhancement has been improved with the development of Anti-Hacking VMAF \cite{ref:ah_vmaf}.

The Enhanced VMAF model (EVMAF) \cite{ref:evmaf} is the most recent attempt at improving the VMAF model. Similar to Ensemble VMAF, EVMAF combines two models, one trained on a private database, and one on a public database. However, EVMAF incorporates motion using a dynamic texture feature, and expands the feature set significantly, using greedy feature selection to obtain the final models. In addition to VMAF v0.6.1, the EVMAF model is used as a baseline against which the performance of FUNQUE is compared.

\section{ALGORITHM}
\label{sec:algorithm}

The foundation of the FUNQUE framework is the use of a unified transform that is shared by all ``atom'' quality models. This unified transform consists of a wavelet transform and HVS processing using a model of the contrast sensitivity function (CSF) and is described in detail in Section \ref{sec:unified_transform}. The following four ``categories'' of atom features have been considered. The feature selection method used to select the final feature set has been described in Section \ref{sec:experiments}. 
\begin{enumerate}
    \item DLM
    \item Structural Similarity (SSIM) \cite{ref:ssim} or Enhanced SSIM (ESSIM) \cite{ref:essim}  computed in the wavelet domain (WD-SSIM and WD-ESSIM), as described in Section \ref{sec:wd_ssim}.
    \item VIF using vector and scalar Gaussian Scale Mixture models \cite{ref:vif}, VIF-Edge and VIF-Approx features \cite{ref:vif_dwt}, and VIF-Scale features, which consist of applying scalar VIF on low-pass subbands at each wavelet level, similar to VMAF's VIF features.
    \item Motion, analogous to VMAF's TD feature, computed as the Mean Absolute Difference between low-pass subbands of successive frames of the reference video.
\end{enumerate}

\subsection{Unified Transform}
\label{sec:unified_transform}

\begin{figure}[t]
    \centering
    \includegraphics[width=\linewidth]{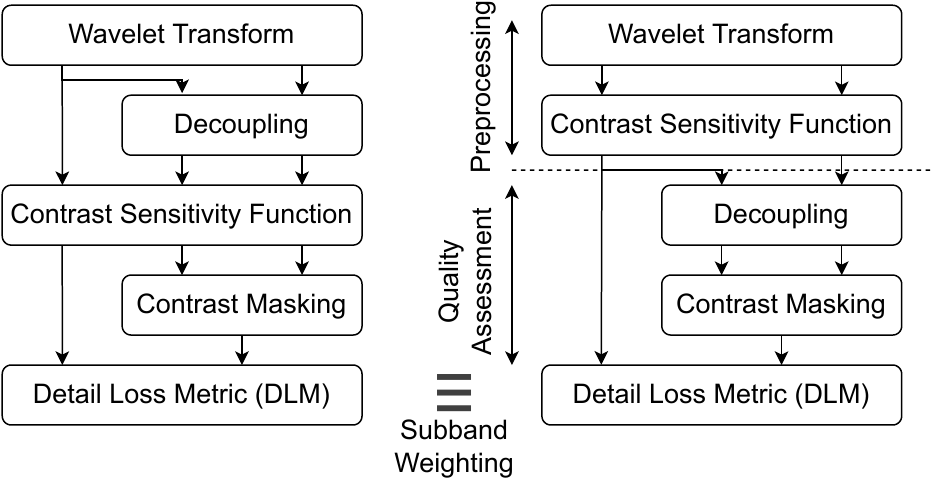}
    \caption{An equivalent reformulation of DLM}
    \label{fig:dlm_flowchart}
\end{figure}

\begin{table}[t]
    \caption{Databases used for model selection}
    \label{tab:databases}
    \begin{center}
        \begin{tabular}{|c|c|c|}
            \hline
            Database &  Size & Codec(s) \\
            \hline
            CC-HDDO \cite{ref:cc_hddo} & 90 & HM, AV1 \\
            BVI-HD \cite{ref:bvi_hd} & 192 & HM \\
            CC-HD \cite{ref:cc_hd} & 108 & HM, AV1, VTM \\
            IVP \cite{ref:ivp} & 100 & Dirac, JM, MPEG-2 \\
            MCL-V \cite{ref:mcl_v} & 96 & x264 \\
            Netflix Public \cite{ref:vmaf} & 70 & x264 \\
            SHVC \cite{ref:shvc} & 64 & HM \\
            VQEGHD3 \cite{ref:vqeghd3} & 72 & MPEG-2, JM \\
            \hline
        \end{tabular}
    \end{center}
\end{table}

The DLM algorithm is illustrated in Figure \ref{fig:dlm_flowchart}. The CSF is applied in DLM by multiplying each subband by a CSF value. We will refer to any such mechanism that assigns weights to each subband as a ``subband-weighting'' (SW) mechanism, and we will refer to the method used in \cite{ref:dlm} as ``Li subband weighting'' (Li SW). On the other hand, the ADM feature used in VMAF is a version of DLM that uses a different set of weights, obtained from \cite{ref:cdf97_watson}. We will refer to this method as ``Watson subband weighting'' (Watson SW).

A closer look at the Decoupling step of DLM reveals that the order of Decoupling and any SW CSF may be reversed. This allows the reinterpretation of DLM's workflow as quality assessment performed on an HVS-aware transform. This raises the possibility of sharing the HVS-aware transform among all the atom features. Indeed, our experiments revealed that CSF sharing leads to a significant boost in performance. Furthermore, since SW is a ``coarse'' CSF method, due to the use of uniform weights within subbands, finer means of applying the CSF may be considered.

The frequency domain is a natural choice for applying the CSF since it is a function of spatial frequency. Consequently, the following frequency-domain model of the CSF was used \cite{ref:csf_frequency}, which was also used to derive Li's subband weights.

\begin{equation}
    \text{CSF}(f) = (0.31 + 0.69f) e^{-0.29f},
    \label{eq:csf}
\end{equation}
where \(f\) has the units cycles/degree. The CSF is applied independently on horizontal and vertical frequencies.

An equivalent ``continuous-angle'' filter in the spatial domain may be obtained using the Inverse Fourier Transform. For practical use, the continuous filter is sampled and truncated to obtain a 21-tap filter that is applied separably in 2D to perform CSF filtering in the spatial domain. The effectiveness of spatial CSF filtering is demonstrated in Section \ref{sec:ablation_results}.

\begin{table}[b]
    \caption{The proposed FUNQUE model}
    \label{tab:funque}
    \begin{center}
        \begin{tabular}{|p{0.45\linewidth}|p{0.14\linewidth}|p{0.1\linewidth}|c|}
            \hline
            Atom Features & Wavelet (Levels) & CSF & SAST \\
             \hline
            WD-ESSIM + VIF-Scales 1 \& 2 + DLM + Motion & Haar (1) & Spatial & Yes \\
            \hline
        \end{tabular}
    \end{center}
\end{table}

\subsection{Wavelet-Domain Structural Similarity}
\label{sec:wd_ssim}

The ESSIM model \cite{ref:essim} improves upon SSIM by using small, strided rectangular windows to compute local quality scores, Coefficient of Variation (CoV) pooling for spatial aggregation, and the Self-Adaptive Scale Transform (SAST) to rescale frames before quality assessment. However, since SAST must be applied commonly to all atom features, its use is investigated as a global design choice in Section \ref{sec:experiments}. So, in this paper, we consider CoV-pooling to be the defining factor of ESSIM, and we consider SSIM to be any mean-pooled version of the algorithm.

In this section, we describe a method for computing SSIM and ESSIM directly from Haar wavelet subband coefficients, which allows us to use the unified transform described in Section \ref{sec:unified_transform}. This is achieved by leveraging the orthonormality of the Haar bases, and the structure of the Haar transform.

Using these properties, local statistics within disjoint blocks of size \(2^L \times 2^L\) may be obtained directly from the wavelet coefficients. For simplicity, consider a pair of images \(x\), \(y\) of size \(M \times N\) such that both \(M\) and \(N\) are divisible by \(2^L\). Now consider their \(L\)-level wavelet decompositions. Let \(H_{x, k}, V_{x, k}, D_{x, k}\), and \(H_{y, k}, V_{y, k}, D_{y, k}\) denote the horizontal, vertical, and diagonal subbands at level \(k\) of their wavelet decompositions respectively, and \(A_{x, L}\), \(A_{y, L}\) be the respective residual low pass (approximation) subbands. Then, local means, variances, and covariances may be obtained as
\begin{equation}
    \mu_{L}(i, j) = 2^{-L}A_L(i, j),
\end{equation}
\begin{equation}
    \sigma^2_{L}(i, j) = 2^{-2L}\sum_{k=1}^{L} \sum_{P^k_{ij}} \sum_{\{H, V, D\}} C_{k}^2(m, n),
\end{equation}
\begin{equation}
    \sigma_{xy, L}(i, j) = 2^{-2L}\sum_{k=1}^{L} \sum_{P^k_{ij}} \sum_{\{H, V, D\}} C_{x, k}(m, n) C_{y, k}(m, n),
\end{equation}
where \(P^k_{ij} = \{(m, n) \mid i2^{L-k} \leq m < (i+1)2^{L-k}, j2^{L-k} \leq n < (j+1)2^{L-k}\}\), and \(C \in \{H, V, D\}\) denotes a subband. These local statistics may be used to compute both WD-SSIM and WD-ESSIM, as in \cite{ref:essim}.
\begin{figure}[t]
    \centering
    \includegraphics[width=0.9\linewidth]{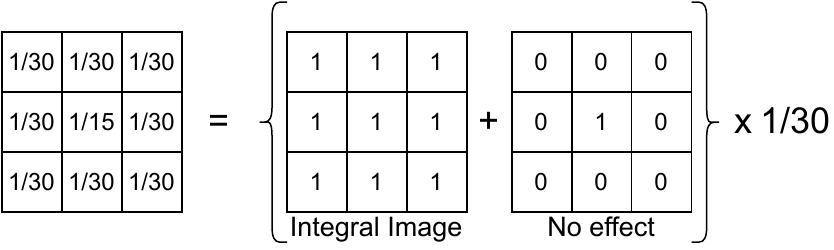}
    \caption{Decomposing DLM's contrast masking kernel}
    \label{fig:dlm_conv_trick}
\end{figure}

\subsection{Integral Image-based Optimization}
\label{sec:algorithm_optimization}
In addition to the use of a unified transform and computing SSIM directly from wavelet coefficients, integral images \cite{ref:viola} have been used to effectively minimize the number of convolution operations. Specifically, we have optimized the computation of local statistics for VIF using integral images, instead of convolution, as in \cite{ref:essim}. Furthermore, the non-separable \(3\times3\) kernel used in DLM has been decomposed as shown in Figure \ref{fig:dlm_conv_trick}. Local sums are then computed using integral images, and the Kronecker delta leaves the image unchanged. Hence, the only convolution in DLM is eliminated.



\section{Experiments}
\label{sec:experiments}

\begin{table*}[t]
    \caption{Comparison of FUNQUE's performance with baseline fusion models}
    \label{tab:funque_performance}
    \begin{center}
        \begin{tabular}{|c|c|c|c|c|c|c|c|c|c|}
            \hline
            Model & BVI-HD & CC-HD & IVP & MCL-V & NFLX-P & SHVC & VQEGHD3 & Average \\
            \hline
            VMAF v0.6.1 & \textbf{0.7962} & 0.8723 & 0.8786 & 0.7766 & 0.9104 & 0.8442 & \textbf{0.9114} & 0.8631 \\
            \textbf{Retrained VMAF v0.6.1} &	0.7516 & \textbf{0.8920} & 0.7156 & 	0.8133 & 	0.8756 & 	0.7205 & 0.7692 & 0.8019 \\
            Enhanced VMAF - M1 &	0.8067 &	0.8595 & 	0.9060 & 	0.8044 & 	0.9168 & 	0.8652 & 	0.9221 & 	0.8761 \\
            \textbf{Enhanced VMAF - M2} &		0.7920 &	0.8376 & 	0.8810 & 	\textbf{0.8327} & 	0.9141 & 	0.8591 & 	0.8729 & 	0.8600 \\
            Enhanced VMAF &	0.8057 &	0.8783 & 	0.9022 & 	0.8282 & 	0.9253 & 	\textbf{0.8796} & 	0.9241 & 	\textbf{0.8842} \\
            \textbf{FUNQUE} & \textbf{0.7959} & 	0.8315 & 	\textbf{0.9186} & 	0.7302 & 	\textbf{0.9358} & 	\textbf{0.8769} &  \textbf{0.9088} & 	\textbf{0.8715} \\
            \hline
        \end{tabular}
    \end{center}
\end{table*}

\begin{table}[b]
    \caption{Timing analysis of FUNQUE models}
    \label{tab:timing_analysis}
    \begin{center}
    \begin{tabular}{|p{0.18\linewidth}|p{0.17\linewidth}|p{0.13\linewidth}|p{0.15\linewidth}|p{0.16\linewidth}|}
    \hline
    Model &	Runtime &	Ops Per Pixel &	Observed Speedup &	Expected Speedup \\
    \hline
    PyVMAF	& 105.23 (s) &	219.61 & 1 & 1 \\
    \textbf{FUNQUE} &	\textbf{12.73 (s)} &	\textbf{39.30} &	\textbf{8.265} &	\textbf{5.588} \\
    \hline
    \end{tabular}
    \end{center}
\end{table}

\begin{figure*}[t]
    \centering
    \includegraphics[width=0.8\linewidth]{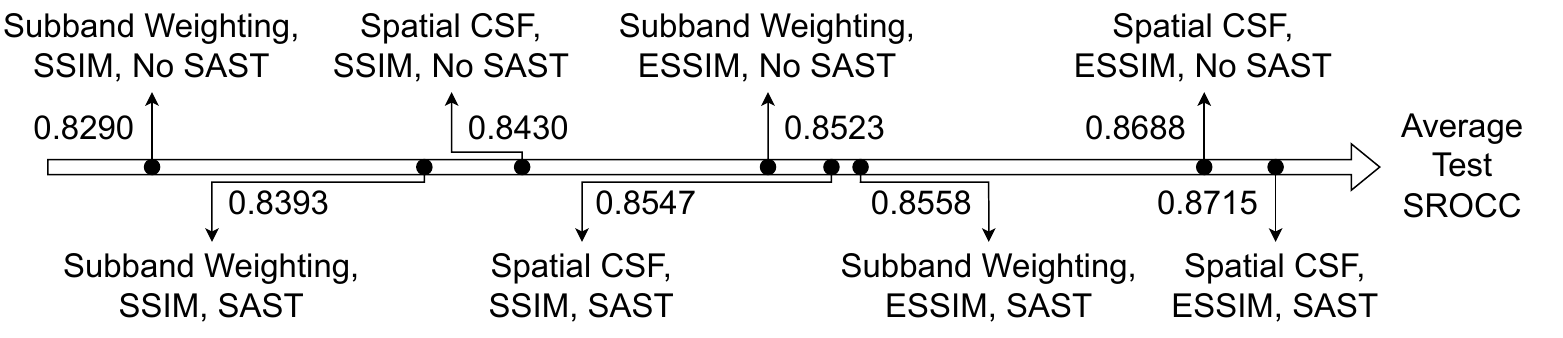}
    \caption{Visualizing the effect of spatial CSF filtering, Enhanced SSIM, and SAST on performance.}
    \label{fig:performance_axis}
\end{figure*}

The framework presented in Section \ref{sec:algorithm} offers a few free design choices. Specifically, let a ``configuration'' correspond to a choice of wavelet, number of wavelet levels, CSF method, whether to share the CSF, and whether to apply SAST. Note that spatial and frequency-domain CSFs must be shared since they are applied before the wavelet decomposition. Table \ref{tab:design_choices} lists all the design choices considered in our experiments, which led to a total of 96 configurations. 

To identify the best configuration, we conducted experiments using the set of 8 databases listed in Table \ref{tab:databases}. These databases were chosen since they represent the popular use-case of video compression, and they were used to develop EVMAF. In order to avoid large models that use several features of the same type, we performed feature selection under the constraint that at most one feature is selected from each category described in Section \ref{sec:algorithm}. An exhaustive search was performed to optimize the Spearman Rank Order Correlation Coefficient (SROCC) during cross-validation over 5000 80-20 splits of the CC-HDDO database. The best model so obtained was tested on the other 7 databases, and the average test SROCC was obtained using Fisher averaging \cite{ref:evmaf}.

\section{RESULTS}
\label{sec:results}

The best FUNQUE model has been described in Table \ref{tab:funque}, and Table \ref{tab:funque_performance} summarizes the correlations against subjective scores achieved by VMAF, EVMAF, and FUNQUE. Since FUNQUE was trained only on the CC-HDDO database, we retrained VMAF v0.6.1 for a fair comparison. Indeed, the retrained model achieved a significantly lower performance, demonstrating the effectiveness of the private database in training. Similarly, we consider EVMAF M2 to be the primary baseline since EVMAF M1, and therefore, the combined EVMAF model, was trained on a private database. The best model on each database among FUNQUE and the ``fair'' baselines has been highlighted in bold. In addition, any better performing ``unfair'' baseline has also been highlighted. 

From Table \ref{tab:funque_performance}, it may be observed that FUNQUE significantly outperforms retrained VMAF v0.6.1, and also outperforms the high-complexity EVMAF M2 model. In addition, despite training only on public data, FUNQUE outperforms VMAF v0.6.1 off-the-shelf and rivals the performance of EVMAF M1, both of which were trained on private Netflix data.

\subsection{Ablation Study}
\label{sec:ablation_results}
In order to understand FUNQUE's superior performance, we would like to investigate three key design choices - the use of ESSIM vs. SSIM, the use of SAST, and the use of the spatial CSF vs. SW CSFs. Since our experiments revealed that not sharing CSF decreased performance significantly, this choice has been omitted from the ablation study.

The performances of the eight models so obtained have been illustrated in Figure \ref{fig:performance_axis}. From the figure, it may be observed that all eight models outperform retrained VMAF v0.6.1, which demonstrates the impact of CSF sharing. Secondly, both ESSIM and Spatial CSF contribute roughly equally to FUNQUE's performance. In other words, despite omitting the expensive 21-tap CSF filter, about 50\% of the reported improvement may be achieved. Finally, we observe that using SAST always improves model performance. Since SAST effectively scales images to half the original resolution, it also improves efficiency.

\subsection{Timing Analysis}

In order to highlight the computational benefits of FUNQUE, estimates of the number of operations per pixel (OPP) required to compute VMAF and FUNQUE were obtained. The ratio of estimated OPPs is reported as the expected speedup. Furthermore, a practical estimate of the speedup was obtained by measuring the ratio of the average running time of the two models on ten videos from the Netflix-Public database.

Since FUNQUE was implemented in Python, the VMAF model was reimplemented in Python for a fair comparison. We refer to this model as PyVMAF. From Table \ref{tab:timing_analysis}, it may be observed that FUNQUE reports a significant speedup of over 8\(\times\) as compared to PyVMAF! Since EVMAF uses more features and requires optical flow estimation, it would already be much slower than VMAF v0.6.1. Therefore, it has not been included in this timing analysis.

\section{CONCLUSION}
In summary, we have proposed a framework to unify quality evaluators by computing them from a common HVS-sensitive transform and fusing them using an SVR. FUNQUE significantly outperforms both the baseline models, i.e., VMAF v0.6.1 and Enhanced VMAF, at less than 1/8th of the computational cost. An open-source implementation of FUNQUE is available at \url{https://github.com/utlive/funque}.

In the future, we see merit in including more sophisticated color and motion-sensitive features, as in ColorVMAF \cite{ref:color_vmaf} and EVMAF. A more extensive feature set may be considered too, with a special focus on wavelet-domain features. Finally, more sophisticated models of the CSF, and even the HVS, may be considered.
\bibliographystyle{IEEEtran}
\bibliography{strings,refs}

\end{document}